# Cognitive Activation and Chaotic Dynamics in Large Language Models: A Quasi-Lyapunov Analysis of Reasoning Mechanisms

Xiaojian Li, Yongkang Leng, Ruiqing Ding, Hangjie Mo*, Shanlin Yang*

School of Management, Hefei University of Technology

**Abstract**

The human-like reasoning capabilities exhibited by Large Language Models (LLMs) challenge the traditional neural network theory's understanding of the flexibility of fixed-parameter systems. This paper proposes the "Cognitive Activation" theory, revealing the essence of LLMs' reasoning mechanisms from the perspective of dynamic systems: the model's reasoning ability stems from a chaotic process of dynamic information extraction in the parameter space. By introducing the Quasi-Lyapunov Exponent (QLE), we quantitatively analyze the chaotic characteristics of the model at different layers. Experiments show that the model's information accumulation follows a nonlinear exponential law, and the Multilayer Perceptron (MLP) accounts for a higher proportion in the final output than the attention mechanism. Further experiments indicate that minor initial value perturbations will have a substantial impact on the model's reasoning ability, confirming the theoretical analysis that large language models are chaotic systems. This research provides a chaos theory framework for the interpretability of LLMs' reasoning and reveals potential pathways for balancing creativity and reliability in model design.

## 1 Introduction

Large Language Models (LLMs) have demonstrated remarkable capabilities in recent years, not only generating fluent natural language text but also performing complex reasoning tasks, solving multi-step problems[1][2], and even exhibiting human-like thinking processes in certain scenarios[3][4][5]. This phenomenon raises a fundamental question: how can these neural network systems, which essentially have fixed parameters, produce seemingly flexible and varied reasoning abilities? Traditional perspectives suggest that large language models merely perform high-dimensional interpolation[6][7] or implicit memorization of training data[8][9], rather than executing genuine reasoning. However, this view struggles to explain these models' excellent performance in zero-shot and few-shot scenarios, as well as their ability to handle complex logic and abstract concepts.

This research proposes a new theoretical framework to understand the reasoning mechanisms of large language models through the phenomenon of "Cognitive Activation." Our core hypothesis is that reasoning abilities in LLMs stem from a dynamic, context-dependent information extraction process, wherein the model dynamically determines how to extract information from its parameters based on current inputs. This process resembles state evolution in dynamical systems and exhibits certain characteristics of chaotic systems.

Traditional neural network research has primarily focused on network architecture design[10][11][12], training methods[13][14], and parameter optimization[15][16], with less exploration of network dynamics during the inference phase. This study approaches from a different perspective, viewing large language models as dynamic systems where each layer's output not only depends on the current input but also determines how subsequent layers extract information from network parameters. This perspective helps explain why models with fixed parameters can exhibit flexible, context-sensitive behaviors.

Chaos theory provides powerful tools for analyzing complex dynamical systems[17]. By introducing quasi-Lyapunov exponents(QLE), we can quantitatively describe the dynamic information extraction process in large language models, revealing the formation mechanism of their reasoning capabilities. This chaos theory-based analytical approach not only helps understand the behavior of existing models but may also provide theoretical guidance for designing more efficient and reliable language models.

The main contributions of this paper include:
- Proposing the concept of "Cognitive Activation" to describe the dynamic information extraction process in large language models.
- Establishing a parameter space correlation analysis framework to elucidate the information flow mechanism between model layers.
- Introducing quasi-Lyapunov exponents to quantitatively analyze model reasoning properties from a chaos theory perspective.
- Proposing a chaos-based functional partitioning method for large language models, distinguishing between convergence and divergence domains.

## 2 Related works

**Neural Network Dynamical Characteristics.** Early studies on the dynamical properties of neural networks laid foundational insights: Funahashi & Nakamura [18]and Rodriguez et al. [19]demonstrated dynamical behaviors in recurrent neural networks (RNNs), though analyses were limited to small-scale architectures. Sussillo & Barak[20] advanced this by characterizing RNN hidden state evolution through low-dimensional manifold extraction, while Haber & Ruthotto [21] framed deep learning as parameter estimation for nonlinear dynamical systems, addressing stability concerns. Saxe et al. [22] uncovered nonlinear learning phenomena in neural networks, including plateau phases during training and abrupt transitions to superior solutions. Chen et al.[23] introduced Neural ODEs to model residual networks as continuous dynamical systems, primarily in vision tasks. Dai et al.[24] reinterpreted self-attention as a higher-order control structure, revealing mixed fast-slow timescale dynamics during forward propagation.

**Chaos Theory in Model Analysis.** Chaos theory provides critical frameworks for analyzing complex system dynamics. Parlitz & Merkwirth[25] pioneered Lyapunov exponent applications for RNN stability, while Liao et al. [26]identified chaotic behaviors in single-neuron delayed systems, showing how nonmonotonic activations and time delays induce Hopf bifurcations and chaotic attractors. Susnjak et al. [27] observed phase transitions in AI systems beyond critical complexity thresholds, marked by irregular oscillations and unpredictability in benchmark performance. Zhang et al. [28] validated the "edge of chaos" hypothesis in LLMs, demonstrating optimal task performance occurs near the ordered-chaotic critical state, aligning with Langton's computational principle [29].

**Transformer Parameter Space Analysis.** Recent work elucidates the structural and functional organization of Transformer parameter spaces. Elhage et al. [30]decoded parameter compositionality mathematically, revealing how components like "induction heads" enable in-context learning via conditional subspace activation. Geva et al. [31] demonstrated that feed-forward layers dynamically update token distributions through vector superposition, with hierarchical "concept neuron" structures. Edelman et al. [32] and Makkuva et al. [33] exposed dual global minima and local traps in loss landscapes through Markov chain tasks, highlighting the interplay between parameter optimization and dynamic information extraction. Park et al. [34] formalized the "linear representation hypothesis" using causal inner products, showing orthogonality for semantically decoupled concepts, while Mickus et al. [35] decomposed embeddings to reveal geometric differentiation by attention and FFN modules. Collectively, these studies establish that parameter spaces employ input-context-driven projections (e.g., attention weights, layer normalization scaling) to selectively activate matrix substructures, governed by nonlinear parameter interactions and dynamic combinatorial rules. Complementary work by Nguyen[36] empirically links this activation mode to training data statistics (e.g., N-gram distributions) and chaotic sensitivity to initial conditions, solidifying the empirical basis for cognitive activation theory.

## 3 Reasoning Paradigm of Large Language Models

After training completion, large language models (LLMs) retain fixed structural elements and parameters. During inference, while these models process new inputs, their fundamental functional properties remain unchanged. This invariance provides an opportunity to analyze the formation mechanism of their reasoning capabilities by meticulously examining how they process and transform input information.

### 3.1 Cognitive Activation Phenomenon

Contemporary LLM architectures are predominantly based on the Transformer framework with sophisticated adaptations[2]. These architectures integrate several critical components: multi-head attention mechanisms for contextual relationship modeling, positional encoding modules that capture sequential information, feed-forward networks for non-linear transformations, layer normalization for training stability, and residual connections that facilitate gradient flow[37].

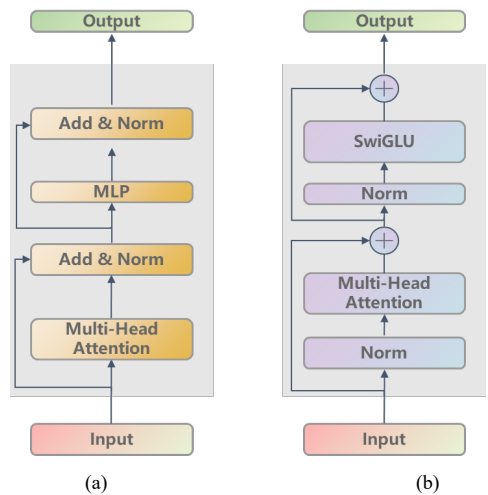

Figure 1. The typical architectures of GPT and LLaMA are depicted. Figure (a) illustrates the architecture based on GPT, while Figure (b) presents the architecture based on LLaMA.

As a typical and mainstream open-source architecture, the LLaMa model demonstrates flexibility and has achieved outstanding results across multiple tasks. Our study use the LLaMa-based architecture[38] as a representative example, we can formally express the forward inference process of a model block through the following mathematical formulation (as established in "Transformer Feed-Forward Layers Build Predictions by Promoting Concepts in the Vocabulary Space"):

$$\begin{cases} X^{(n+1)} = f_{MLP}^{(n)}\left(\text{Norm}(X^{(n)'})\right) + X^{(n)'} \\ X^{(n)'} = f_{att}^{(n)}\left(\text{Norm}(X^{(n)})\right) + X^{(n)} \end{cases} \quad (1)$$

Where:
- $X^{(n)}$ is the feature matrix (hidden states) at layer $n$ of the model, representing the transformed input as it passes through the $n$-th layer of the network.
- $X^{(n)'}$ is an intermediate feature matrix after the self-attention mechanism but before the norm of feed-forward network in layer $n$, representing the output after applying attention and adding the residual connection.
- $X^{(n+1)}$ is the feature matrix output from layer $n$, which becomes the input to layer $n + 1$.
- $\text{Norm}(\cdot)$ is the layer normalization function that stabilizes the values of the feature vectors by normalizing their mean and variance.
- $f_{att}^{(n)}(\cdot)$ is the multi-head self-attention function at layer $n$, defined as:

$$f_{att}^{(n)}(X) = \oplus_{j=1}^{H} \left( \text{softmax}\left( \frac{\left(XW_{Q_j}^{(n)}\right)\left(XW_{K_j}^{(n)}\right)^T}{\sqrt{\frac{d}{H}}} \right) \left(XW_{V_j}^{(n)}\right) \right) W_{O_j}^{(n)} \quad (2)$$

Where:
- $H$ is the number of attention heads
- $W_{Q_j}^{(n)}$ is the weight matrix for queries in the $j$-th attention head at layer $n$
- $W_{K_j}^{(n)}$ is the weight matrix for keys in the $j$-th attention head at layer $n$
- $W_{V_j}^{(n)}$ is the weight matrix for values in the $j$-th attention head at layer $n$
- $W_{O_j}^{(n)}$ is the output projection matrix for the $j$-th attention head at layer $n$
- $d$ is the dimensionality of the model's hidden state
- $\sqrt{d/H}$ is the scaling factor to prevent extremely small gradients during training

- $f_{MLP}^{(n)}(\cdot)$ is the multi-layer perceptron (feed-forward network) function at layer $n$, defined as:

$$f_{MLP}^{(n)}(X) = g\left(XW_1^{(n)}\right) W_2^{(n)} \quad (3)$$

Where:
- $W_1^{(n)}$ is the weight matrix for the first linear transformation in the feed-forward network at layer $n$
- $W_2^{(n)}$ is the weight matrix for the second linear transformation in the feed-forward network at layer $n$
- $g(\cdot)$ is the non-linear activation function (typically GELU or ReLU).

Equation (1) reveals a critical insight into the deterministic nature of these models: the output of layer n+1 is strictly dependent on the output of layer n. With constant model parameters, this output is exclusively determined by the preceding layer's output. Consequently, from a functional perspective, in the absence of stochastic elements, the model's final output is entirely determined by its initial input. This observation has led many researchers to conclude that LLMs merely implement implicit encoding and high-dimensional interpolation of existing knowledge systems, without possessing genuine analytical or reasoning capabilities.

However, we propose a paradigm shift in conceptualizing the LLM reasoning process. By inverting our perspective—treating model inputs as function parameters and model parameters as function variables—equation (1) can be reformulated as:

$$\begin{cases} X^{(n+1)} = f_{MLP}^{(n)}\left(W_1^{(n)}, W_2^{(n)}\right) + X^{(n)'} \\ X^{(n)'} = f_{att}^{(n)}\left(W_{Q_j}^{(n)}, W_{K_j}^{(n)}, W_{V_j}^{(n)}, W_{O_j}^{(n)}\right) + X^{(n)} \end{cases} \quad (4)$$

In this alternative formulation, the expressions for $f_{MLP}^{(n)}\left(W_1^{(n)}, W_2^{(n)}\right)$ and $f_{att}^{(n)}\left(W_{Q_j}^{(n)}, W_{K_j}^{(n)}, W_{V_j}^{(n)}, W_{O_j}^{(n)}\right)$ remain functionally identical to those of $f_{MLP}^{(n)}(\text{Norm}(X^{(n)'}))$ and $f_{att}^{(n)}(\text{Norm}(X^{(n)}))$, respectively. However, this reframing reveals a crucial insight: with a fixed neural network architecture, the functional properties of $f_{MLP}^{(n)}$ and $f_{att}^{(n)}$ are dynamically determined by the input information $X^{(n)}$ and $X^{(n)'}$. The scientific consensus holds that world knowledge acquired by neural networks is implicitly encoded within their parameter matrices—specifically within $W_1^{(n)}, W_2^{(n)}, W_{Q_j}^{(n)}, W_{K_j}^{(n)}, W_{V_j}^{(n)}, W_{O_j}^{(n)}$—which remain invariant during inference. Therefore, when input information $X^{(n)}$ changes, the function properties of $f_{MLP}^{(n)}$ and $f_{att}^{(n)}$ transform correspondingly. This transformation alters the parameter processing methodology, which subsequently modifies both the model output $X^{(n+1)}$ and the specific information extracted from the implicit encoding. This insight leads us to formulate the following theorem.

***Theorem 1***: In large language models, for any neural network module with fixed structure, we can express it as a functional mapping:

$$f_{X_n}: \mathbb{W} \to \mathbb{X}$$

where $\mathbb{W}$ represents the parameter space of the neural network module, and $\mathbb{X}$ denotes the feature space of neural network outputs. The functional properties of this mapping are uniquely determined by the input matrix $X^{(n)}$.

Theorem 1 encapsulates a fundamental principle: the input to a neural network deterministically shapes the pattern by which information is extracted from neural network parameters. Building upon this theorem, we introduce the central definition of our theory:

***Definition 1 (Cognitive Activation Phenomenon)***: The cognitive activation phenomenon refers to the process whereby the pattern of information extraction from neural network parameters is dynamically determined by outputs generated from preceding neural network inference operations.

In contemporary LLM implementations, cognitive activation manifests across three distinct hierarchical levels that form an integrated system of information processing:

(1) Intra-network Cognitive Activation: During forward inference in LLMs, the output from

one block determines the information extraction strategy employed by subsequent blocks. This creates a cascade of information processing where each layer's computation is contextually influenced by preceding computations.

(2) Iterative Cognitive Activation: In the next-token prediction process, each generated token becomes integrated into the subsequent input context, thereby influencing the information extraction strategy for generating future tokens. This creates a feedback loop where model outputs recursively shape future processing.

(3) Reasoning Cognitive Activation: In models incorporating chain-of-thought reasoning (e.g., GPT-o1, DeepSeek-R1), the complete text output from a previous reasoning step determines the thinking strategy for subsequent steps, thereby shaping the information extraction methodology at a higher cognitive level.

Additionally, the attention mechanism itself exhibits analogous characteristics within this framework—matrices Q and K, processed through $W_{Q_j}^{(n)}$ and $W_{K_j}^{(n)}$, determine the information extraction strategy from parameter matrix $W_{V_j}^{(n)}$, creating micro-level cognitive activation patterns within individual computational units.

The essential characteristic of cognitive activation lies in its interconnected and dynamic nature of information extraction. This interconnectedness constitutes the fundamental rationale for terming this phenomenon "cognitive activation"—it captures how specific inputs "activate" particular information patterns within the model's parameter space. We posit that in well-trained LLMs demonstrating reasoning capabilities, these abilities emerge precisely from this interconnected processing mechanism.

The cognitive activation process creates a sophisticated cycle: each module's input activates specific information encoded within model parameters, generating an output that simultaneously determines the input for subsequent modules. Through this recursive cycle, LLMs develop emergent reasoning capabilities that bear resemblance to human intelligence, despite operating with fixed parameters. This perspective helps reconcile the apparent paradox of how static parameter matrices can produce dynamic, context-sensitive, and apparently reasoning-based outputs.

**3.2 Parameter Space Correlation**

The cognitive activation phenomenon enables neural networks to dynamically define their information extraction strategies during inference. This makes the formation process of these strategies critically important for understanding how LLMs develop reasoning capabilities. Using the LlaMa-based block as our analytical framework, we can observe how information extraction strategies emerge through the network's computational processes. This allows us to express the m-th row vector of the feature matrix as follows:

$$\begin{cases} x_m^{(n+1)} = mlp_m^{(n)} + x_m^{(n)'} \\ x_m^{(n)'} = att_m^{(n)} + x_m^{(n)} \end{cases} \quad (5)$$

Where:
- $x_m^{(n)}$ is the $m$-th row vector of feature matrix $X^{(n)}$
- $x_m^{(n)'}$ is the $m$-th row vector of intermediate feature matrix $X^{(n)'}$

- $x_m^{(n+1)}$ is the $m$-th row vector of feature matrix $X^{(n+1)}$
- $mlp_{m,n}$ represents the contribution from the feed-forward network, defined as:
$$mlp_m^{(n)} = i_m f_{MLP}^{(n)}\left(W_1^{(n)}, W_2^{(n)}\right)$$
- $att_{m,n}$ represents the contribution from the attention mechanism, defined as:
$$att_m^{(n)} = i_m f_{att}^{(n)}\left(W_{Q_j}^{(n)}, W_{K_j}^{(n)}, W_{V_j}^{(n)}, W_{O_j}^{(n)}\right)$$
- $i_m$ is a row selection vector that extracts the $m$-th row from a matrix, expressed as:
$$i_m = [0,0,\ldots,0,1,0,\ldots,0]$$
where the value 1 appears only in the $m$-th position

From equation (5), we can derive a more comprehensive expression that captures how information flows through the network:

$$x_m^{(n+1)} = mlp_m^{(n)} + att_m^{(n)} + x_m^{(n)} = \sum_{p=0}^{n}\left(mlp_m^{(p)} + att_m^{(p)}\right) + x_m^{(0)} \qquad (6)$$

Equation (6) reveals a profound insight: the input $x_m^{(n+1)}$ to layer $n+1$ of the LLM can be expressed as a linear sum of the outputs from all preceding MLP modules $(mlp_m^{(p)})$, attention modules $(att_m^{(p)})$, and the initial input $(x_m^{(0)})$.

By extension, the final output of the large language model can also be represented as a linear weighted combination of all module outputs and the initial input. This formulation demonstrates that information from earlier layers remains accessible throughout the network, in other words, the final output of a large language model is collectively determined by the intermediate outputs of all preceding layers. This process of evolution in the parameter space can be described from two dimensions: the adjustment of the output's magnitude and the adjustment of the output's direction. The change in magnitude reflects the accumulation of information strength during transmission, while the change in direction embodies the semantic reconstruction of information during transmission. The interplay of these two factors ultimately determines the model's output result.

In order to further reveal the information extraction process of the initial input in the parameter space, we designed a set of experiments aimed at analyzing the variation trend of the intermediate layer output magnitude with respect to the number of layers[1].

The experimental results, as shown in Figure 2(a), indicate that the magnitude exhibits a significant piecewise linear characteristic in the logarithmic coordinate system (with base e), meaning that the actual growth follows an exponential pattern: the linear fitting coefficient for layers 0-9 is 0.27, corresponding to an exponential growth factor of 1.32 in the real domain; for layers 10-38, the linear coefficient decreases to 0.075, and the actual exponential growth factor is 1.08 (the fitting effect is illustrated in Figure 2(c)). This suggests that the information extraction pattern in the parameter space is not uniformly accumulative but rather exhibits a characteristic of accelerated expansion.

Specifically, Figure 2(b) presents the evolution curves of the magnitude ratio for each token. It is observable that a minority of tokens (for example token 0 and token 1) exhibit a step-like pattern at specific layers. In fact, these tokens are part of the system input provided by the model, while for the user-provided inputs, they all demonstrate the aforementioned logarithmic linear growth pattern.

---

[1] The experiments in this paper are conducted based on the open-source model Qwen2-14B[39], which is built upon the LlaMa architecture, consistent with the derivations in the preceding text. This general-purpose architecture, pretrained on an extensive corpus of 3 trillion tokens encompassing diverse data domains, features 40 transformer layers and employs a hidden dimension of 5120.

Further analysis of inter-layer correlations readily discloses that, due to the overall asymptotic trend in processing information, the closer the layers are, the stronger their correlation. Figure 2(d) shows the variation of the standard deviation of magnitude with cross-layer intervals; when the cross-layer interval reaches 6 layers, its standard deviation expands to twice that of adjacent layers. For a more intuitive observation of this phenomenon, we visualize in Figure 3(a) the projection of feature vectors at specific layers (with the final output as the benchmark, the horizontal axis representing the corresponding magnitude ratio and the vertical axis representing the corresponding angle). The results indicate that feature vectors within the same layer exhibit similar patterns, and closer layers have more proximate magnitudes and angles in space. Figure 3(b) employs the Pearson correlation coefficient to calculate the correlation between layers, where adjacent layers show higher correlations, which decay with layer distance. This figure also effectively demonstrates the hierarchical pattern of the model: the correlations between intermediate layers decay more slowly, maintaining a high similarity pattern even over larger cross-layer intervals, whereas the initial layers (before the 10th layer) and the final layers (the last 5 layers) exhibit stronger "independence."

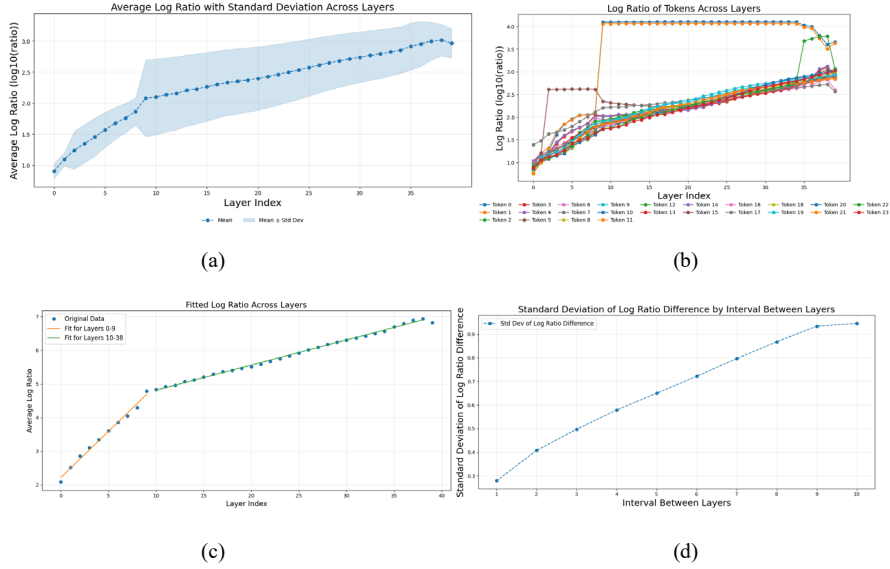

Figure 2. The experimental methodology is as follows: (i) The input encoding $h^{(0)}$ was normalized as $h^{(0)} = \frac{h^{(0)}}{\|h^{(0)}\|_2}$ to ensure uniformity in initial conditions across all samples, thereby isolating the internal geometric dynamics of the network. (ii) For each layer, the modulus ratio $R_i^{(l)} = \frac{\|h_i^{(l)}\|_2}{\|h_i^{(0)}\|_2}$ (where $i=1, \ldots, N$, and $N$ is the number of token vectors after encoding) was computed for the activation vectors relative to the normalized input. (iii) To capture the exponential growth characteristics of hierarchical feature intensity, the logarithmic average of the modulus ratio $R^{(l)} = \frac{1}{N}\sum_i^N \log R_i^{(l)}$ was calculated, serving as a metric for the feature strength at each layer. We conducted multiple input experiments, and all inputs exhibited the same pattern. Here, we use the input "Cats are animals" as an example for presentation. Figure (a) shows the average error across all token vectors, while (b) illustrates the trend of each individual token vector, (c) illustrates the linear fitting segmented into two parts, (d) presents the inter-layer variance at varying degrees of cross-layer intervals.

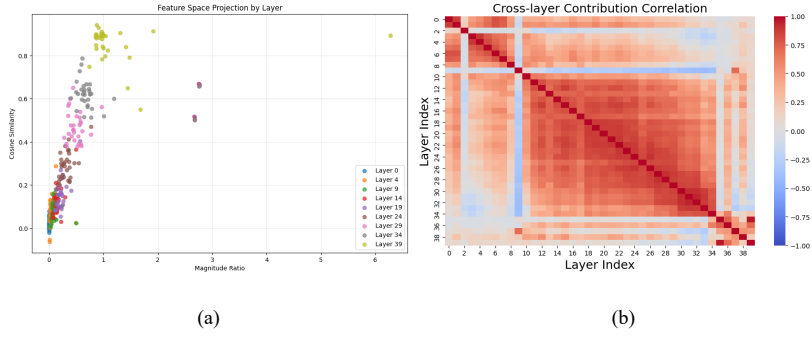

(a)                  (b)

Figure 3. (a) Visualization of the projections of feature vectors at specific layers (with the final output as the benchmark, the horizontal axis represents the corresponding magnitude ratio, and the vertical axis represents the corresponding angle), (b) Calculation of the inter-layer correlations using the Pearson correlation coefficient.

To observe the influence of each component on the output at every layer, we further visualize the evolution trends of the magnitude and angle of the output results from the mlp and attention mechanisms at each layer, as shown in Figure 4. It can be seen that, prior to the final layer, the outputs of both mlp and attention evolve in the same direction as the final output (cosine similarity greater than 0), but the attention in the final layer acts in the opposite direction (cosine similarity less than 0), which also explains the declining trend in Figure 2(a) at the last layer. The step change at the 9th layer mentioned earlier is caused by a sharp increase in the magnitude of the mlp output. It is evident that, in most cases, both the magnitude and cosine similarity of the mlp are higher than those of attention, suggesting that the mlp output plays a dominant role in the final result.

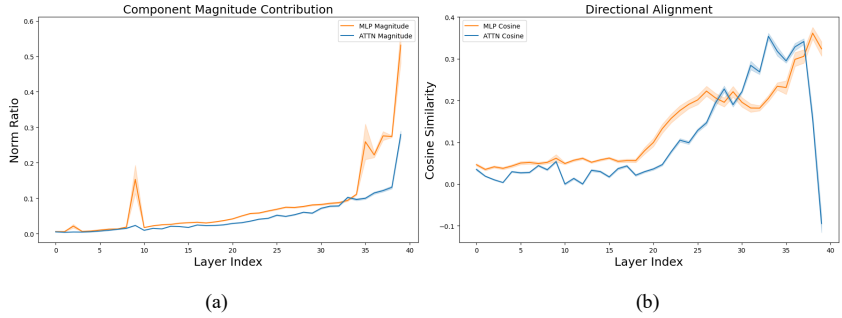

(a)                  (b)

Figure 4. (a) and (b) represent the magnitude ratio and angle between the outputs of the mlp and attention at each layer and the final output, respectively.

To validate the aforementioned points, we decompose the final output result ($x_m^{(n+1)}$) based on the projections of each component ($mlp_m^{(p)}, att_m^{(p)}, x_m^{(0)}$) onto it, with the results depicted in Figure 5. In the final output, 55.7669% is determined by the projection of the mlp output, 44.2322% by the projection of the attn output, while the initial input, after multiple iterations, accounts for only 0.0009% of the final output. This experiment not only verifies that the final output is the sum of the outputs from the mlp, attention, and the initial input, but also illustrates that the mlp plays a more significant role in the final output. The initial input, after undergoing multiple iterations, occupies a negligible proportion in the final output, and we can even approximate it as a minor perturbation to the network!

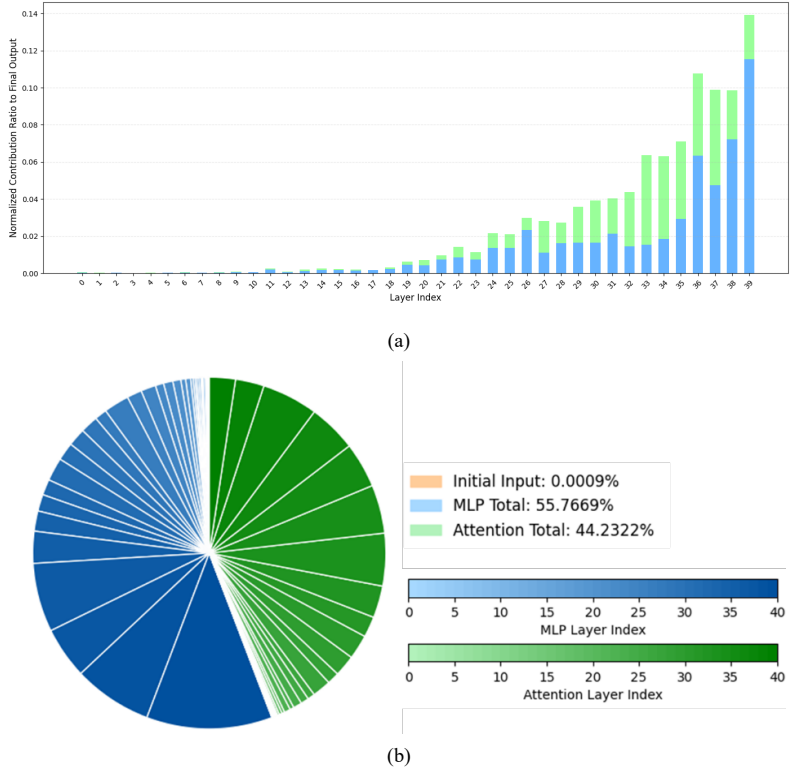

(a)

(b)

Figure 5. Visualization of the projections of the mlp and attention at each layer onto the final result, where (a) allows for the observation of the variation trends of the outputs from different layers, and (b) presents the overall proportion of each component.

## 4 Large Model Reasoning Properties from a Chaos Perspective

### 4.1 Quasi-Lyapunov Exponents of Large Language Models

Chaos theory provides powerful analytical tools for understanding complex dynamical systems. The maximal Lyapunov exponent, in particular, has been widely used to characterize chaotic behavior in various systems. According to established literature in chaos theory, the maximal Lyapunov exponent can be mathematically expressed as:

$$\lambda = \lim_{t\to\infty}\lim_{|\delta_0|\to 0} \frac{1}{t}\ln\frac{|\delta_t|}{|\delta_0|} \tag{7}$$

where $\delta_0$ represents an initial infinitesimal perturbation to the system state, $\delta_t$ represents how this perturbation evolves after time $t$. The ratio $\frac{|\delta_t|}{|\delta_0|}$ quantifies the divergence or convergence of initially close trajectories, $\lambda$ measures the average exponential rate of separation of infinitesimally close trajectories.

For discrete-time systems described by the mapping $x_{n+1} = f(x_n)$, with an orbit starting at $x_0$:

$$\lambda(x_0) = \lim_{n\to\infty}\frac{1}{n}\sum_{i=1}^{n-1}\ln|f'(x_i)| \tag{8}$$

We propose that Lyapunov exponent concepts can be effectively adapted to analyze cognitive activation phenomena in LLMs. However, LLMs differ from traditional dynamical systems in several important respects, necessitating adjustments to the classical definition of Lyapunov exponents. In our analysis, we focus on intra-network and iterative cognitive activation, examining their chaotic characteristics separately.

The third type—reasoning cognitive activation—presents additional complexities due to the semantic richness of large text segments and the rule-governed nature of chain-of-thought mechanisms. These complexities make quantitative analysis particularly challenging and will be addressed in future research.

**4.2 Intra-network Cognitive Activation**

For intra-network cognitive activation, we conceptualize the system state $x_n$ as the feature matrix $x_{m,n+1}$ at each layer of the LLM. The system model corresponds to the LlaMa block at each layer, and the time parameter $n$ aligns with the layer number in the network architecture.

Critical differences from classical dynamical systems must be acknowledged: 1. In LLMs, each layer's LlaMa-block exhibits distinct functional characteristics, unlike the uniform mapping function $f(x_n)$ in standard discrete systems 2. While time in discrete systems can approach infinity, LLMs have a finite, predetermined number of layers 3. The high-dimensional nature of LLM feature spaces requires special consideration for perturbation analysis

These differences necessitate a modified approach to Lyapunov exponents. We therefore introduce a quasi-Lyapunov exponent specifically designed to analyze intra-network cognitive activation in LLMs:

***Definition 2 (Quasi-Lyapunov Exponent for Intra-network Cognitive Activation)***: For any given input embedding $X_0$, if the output of the $m$-th block of the large model is $X_m$, then its quasi-Lyapunov exponent can be defined as:

$$\lambda_{m,n}(X_m) = \lim_{|\delta_m| \to 0} \frac{1}{n-m} \ln \frac{|\delta_{m,n}|}{|\delta_m|} \qquad (9)$$

Where: $\delta_{m,n} = X'_n - X_n = f_{m,n}(X_m + \delta_m) - f_{m,n}(X_m)$ represents the difference in network outputs at layer $n$ resulting from a perturbation $\delta_m$ at layer $m$, $X_m = f_{0,m}(X_0)$ represents the output of the $m$-th block when processing the initial input $X_0$. $f_{m,n}$ denotes the composite function representing the neural network transformation from block $m$ to block $n$. This quasi-Lyapunov exponent quantifies how sensitive the network is to perturbations introduced at intermediate layers, providing insight into how information extraction patterns evolve through the network depth.

In practice, due to the varying activation intensities across different layers, using absolute values may result in neglecting the inherent magnitudes of the elements. Therefore, we can employ dynamically adapted perturbations, specifically setting the percentage of the elements as the perturbation, and refine the approach as follows:

$$\lambda_{m,n}(X_m)' = \lim_{|k| \to 0} \frac{1}{n-m} \ln \frac{|\delta'_{m,n}|}{|\delta'_m|} \qquad (10)$$

Where: $\delta'_m = kX_m$, is a perturbation that dynamically adjusts according to the magnitude of the element itself, $\delta'_{m,n} = f_{m,n}(X_m + \delta'_m) - f_{m,n}(X_m)$, and other representations are as per the aforementioned Equation 9.

While LLMs' reasoning capabilities are often attributed to their static knowledge

encoding[40][41][42], whether their dynamic reasoning processes exhibit chaotic-like sensitivity to initial conditions remains unexplored. Traditional metrics (e.g., accuracy, perplexity) and robustness tests (e.g., adversarial attacks) focus on final outputs, ignoring perturbation propagation through intermediate layers. Guided by cognitive activation theory, this study introduces QLE to quantify dynamic sensitivity, examining whether perturbations in intermediate layers propagate divergently (amplifying semantic shifts) or convergently (stabilizing syntactic structures), thereby revealing the chaotic dynamics underlying LLM reasoning.

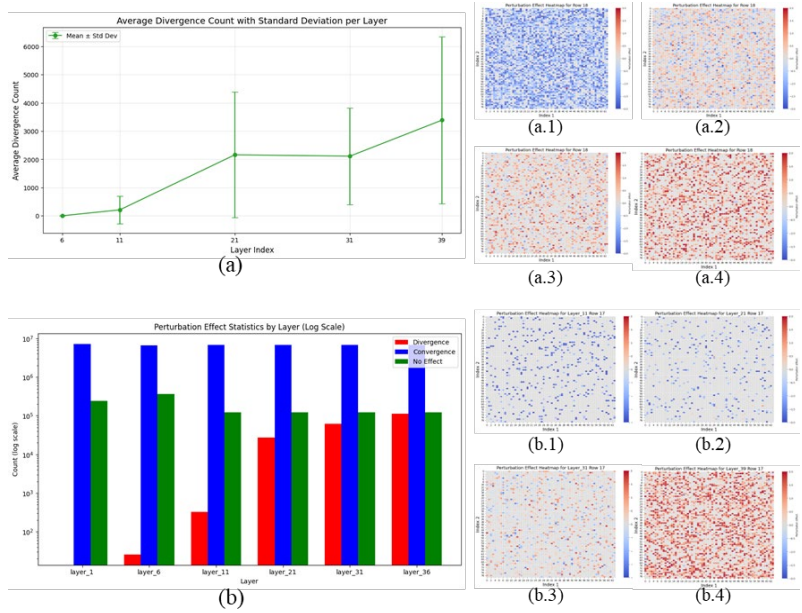

Figure 6. To validate the hypothesis of inter-layer dynamic sensitivity according to the cognitive activation theory, we introduced controlled micro-perturbations to the intermediate layer activations of the model and quantified the propagation of these perturbations using QLE. The specific procedure is as follows: (i) Let the output activation tensor of the intermediate layer $l$ be $H^{(l)} \in \mathbb{R}^{n \times d}$. Select the $k$-th row vector $h_k^{(l)} \in \mathbb{R}^d$. Apply a perturbation $\delta$ or $\delta'$ to each element $h_{k,j}^{(l)}$ of this vector, generating the perturbed vector $\widetilde{h_k^{(l)}} = h_k^{(l)} + \delta \cdot e_j$, where $e_j$ is the $j$-th standard basis vector. (ii) Replace $h_k^{(l)}$ with the perturbed activation vector $\widetilde{h_k^{(l)}}$ to form the modified tensor $\widetilde{H^{(l)}}$, which is then used as the input for layer $l$. Record the resulting output tensor $\widetilde{H^{(l+1)}}$ and compare it with the baseline output $H^{(l+1)}$ (generated without perturbation, with a fixed random seed). Compute the element-wise difference $\Delta_{i,j} = \widetilde{H^{(l+1)}_{i,j}} - H^{(l)}_{i,j}$. (iii) The convergence or divergence effect at each position is calculated using $\lambda_{i,j} = \lim_{|\delta| \to 0} \ln \frac{|\Delta_{i,j}|}{|\delta|}$. If $\lambda_{i,j} > 0$, the node is labeled as divergent; otherwise, it is labeled as convergent. Nodes are classified as divergent or convergent based on the calculated QLE. Using the input "Cats are animals" as an example, we demonstrate the convergence or divergence effects of a minor perturbation at a specific point in the upper layers (here, we fixed on the special position $h_{17,4074}^{(l)}$ for perturbation 0.01 and $h_{17,3500}^{(l)}$ for perturbation $1\% X_{m,j}$, $l = 10, 20, 30, 39$) on the subsequent lower layers.

The QLE for the subsequent layer were calculated, and a heatmap of each position was visualized in figure 6. The results reveal that minor perturbations can induce divergent effects in the reasoning process, with these effects exhibiting a strong dependence on layer depth. Shallow layers predominantly respond to perturbations with convergence, while deeper layers demonstrate pronounced divergence. This indicates the presence of chaotic regions in the deeper layers, which play a crucial role in shaping the information extraction patterns of the model. Such hierarchical chaotic characteristics reveal the model's dual capabilities: the shallow layers extract stable semantic features through convergence mechanisms, while the deeper layers achieve context-driven information reorganization through chaotic sensitivity.

### 4.3 Iterative Cognitive Activation

For iterative cognitive activation, the system state is not characterized by the internal feature representations but rather by the text inputs generated during each iteration. During inference, LLMs select tokens for output based on computed feature vectors, effectively discretizing the parameter space and making infinitesimal perturbations analytically challenging. To address this challenge, we use the model's input embedding matrix as the system state, the entire neural network as the system model, and the number of model iterations as the time parameter. This formulation approximates a conventional dynamical system with consistent model characteristics during iteration, though with a finite time horizon bounded by the text generation process. We therefore propose a quasi-Lyapunov exponent tailored for analyzing iterative cognitive activation:

***Definition 3 (Quasi-Lyapunov Exponent for Iterative Cognitive Activation)***: For any given initial input embedding $X_0$, if the text obtained after $m$ iterations of the large model, after input embedding, is $X_m$, then its quasi-Lyapunov exponent can be defined as:

$$\lambda_m(X_0) = \lim_{|\delta_0| \to 0} \frac{1}{m} \ln \frac{|\delta_m|}{|\delta_0|} \qquad (11)$$

where $\delta_m = X'_m - X_m$ represents the difference in embedded representations after $m$ iterations, $X_{n+1} = N(X_n)$ represents the iteration process, with $N$ being the function representing the entire neural network, $X'_m$ denotes the input embedding matrix obtained after $m$ iterations when using $X_0 + \delta_0$ as the initial input.

This formulation allows us to analyze how small perturbations in initial conditions propagate through multiple inference iterations, capturing the dynamic nature of text generation in LLMs.

### 4.4 Analysis of Large Model Reasoning Properties under Initial Value Perturbation

The quasi-Lyapunov exponents introduced above provide powerful analytical tools for understanding the emergent reasoning properties of LLMs. By examining how perturbations propagate through the network—both within a single forward pass and across multiple generation iterations—we can characterize different regions of the model's operational parameter space.

When $\lambda_{m,n}(X_m) > 0$, the model exhibits sensitivity to initial conditions characteristic of chaotic systems. Small perturbations in the feature representation at layer $m$ lead to significantly different representations at layer $n$. This sensitivity indicates that the model extracts fundamentally different information patterns from its parameter matrices when processing slightly different inputs. Such regions of parameter space may correspond to the model's capacity for divergent thinking, creativity, and context-sensitive reasoning.

Conversely, when $\lambda_{m,n}(X_m) < 0$, the model demonstrates convergent properties. Different inputs within a neighborhood of parameter space produce similar outputs, suggesting that the model has learned to map diverse but related inputs to similar semantic representations. These regions may correspond to the model's capacity for generalization, abstraction, and recognition of underlying patterns despite surface-level differences.

When $\lambda_{m,n}(X_m) \approx 0$, the model operates in a regime where inputs and outputs maintain approximately one-to-one correspondence. In these regions, the model functions primarily as a transformation system, preserving the informational structure of the input while potentially recording it into different representational formats.

For a complete characterization of an LLM's reasoning properties, we must consider both intra-network and iterative quasi-Lyapunov exponents. A model for which both $\lambda_{m,n}(X_m) \leq 0$ and $\lambda_m(X_0) \leq 0$ across all relevant inputs would demonstrate global convergence properties, with different initial conditions ultimately leading to similar outputs. While such convergence may be desirable for specific tasks requiring robust generalization, it would limit the model's capacity for generating diverse and context-sensitive responses.

The interplay between convergent and divergent regions in parameter space likely underlies the sophisticated reasoning capabilities observed in state-of-the-art LLMs. Through the framework of cognitive activation and chaos theory, we gain insight into how fixed-parameter neural networks can nonetheless exhibit flexible, context-sensitive information processing that resembles human reasoning.

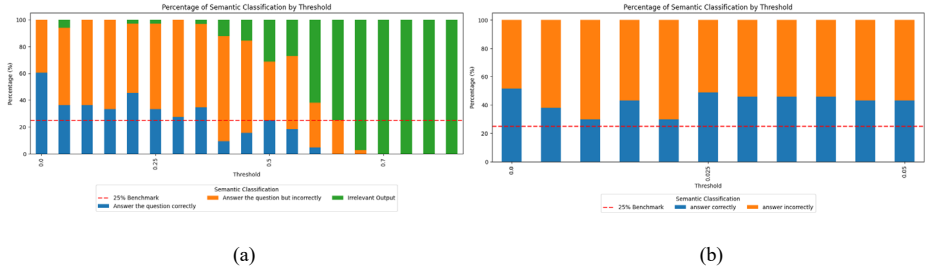

(a)　　　　　　　　　　　　　　(b)

Figure 7. Building upon the hypothesis of dynamic information flow in the parameter space as outlined in the cognitive activation theory, we simulated perturbations in initial conditions by suppressing low-activation neurons. Specifically, during the forward propagation process, for each layer's output tensor $H^{(l)}$, the elements were sorted by their absolute values, and the lowest $k\%$ of elements were set to zero. To better quantify the model's capabilities and minimize the influence of subjective human factors, the experiment randomly sampled 11,528 multiple-choice questions from the CMMLU Bench[44], covering 67 topics. The outputs were categorized into three types: answer the questions correctly, answer the questions but incorrectly, and irrelevant outputs, and were independently evaluated by two annotators.

The changes in reasoning capabilities under minor perturbations to the initial conditions are shown: the left figure illustrates the results with a step size of 5%, while the right figure further refines the analysis with a step size of 0.5‰.

The cognitive activation theory posits that the reasoning capabilities of large language models stem from the dynamic extraction of information within the parameter space, and figure 6 provides preliminary evidence that their behavior resembles that of chaotic systems. According to this theory, the model's reasoning outputs should exhibit sensitivity to initial conditions, manifesting typical characteristics of chaotic systems. Therefore, it is imperative to unveil the chaotic characteristics by quantitatively analyzing the model's behavior under perturbations of initial conditions.

As can be inferred from the results of Experiment 7, when only the 5% of neurons with the smallest absolute values were zeroed out, the model's accuracy in answering questions dropped by over 20%. Further reducing the zeroing proportion in increments of 0.5% resulted in a nearly 14% decline in accuracy. Unlike studies on adversarial attacks, this experiment focuses on system sensitivity under benign perturbations rather than malicious sample attacks. The results reveal that even minor, non-adversarial perturbations can significantly undermine the reliability of the reasoning process. These findings indicate that the model's behavior exhibits sensitivity to initial conditions, showcasing characteristics typical of chaotic systems.

## 5 Discussion

This paper reveals the dynamic nature of reasoning capabilities in LLMs through cognitive activation theory and proposes quantitative analytical tools based on the framework of chaos theory. Experimental results demonstrate that the information extraction process of the model exhibits significant chaotic characteristics, and the formation of its reasoning abilities is closely related to the dynamic evolution of the parameter space.

In the reasoning process of LLMs, the information flow in the parameter space can be regarded as an evolutionary trajectory under a chaotic dynamical system. The three-phase dynamics suggest that the model's information processing may follow an "attractor-guided" mechanism: the initial layers approach fixed-point attractors, the deeper layers tend toward strange attractors, and finally, stability is maintained through compression adjustments. This hierarchical attractor pattern may explain how the model achieves dynamic balance with fixed parameters—different inputs are mapped to specific attractor domains through cognitive activation mechanisms, thereby generating diverse outputs while preserving semantic consistency. This conjecture awaits further validation.

Chaotic characteristics (e.g., sensitivity, bifurcation) directly influence the model's reasoning robustness and creativity. The analysis of QLE shows that deeper layers exhibit significant divergence in response to minor perturbations, while shallower layers are primarily convergent. The performance decline caused by initial condition perturbations is highly consistent with the sensitivity to initial values in chaotic systems, indicating that the model's reasoning process inherently relies on the dynamic equilibrium of the parameter space. However, this sensitivity also poses potential risks—slight benign perturbations may trigger an "avalanche effect", leading to broken reasoning chains or factual errors.

The phenomenon of machine hallucinations can be reinterpreted as a consequence of chaotic dynamics. In the iterative cognitive activation process of LLMs (e.g., chain-of-thought), the model continuously reconstructs the input context through self-feedback mechanisms. If initial perturbations or instabilities in the parameter space are not effectively suppressed, minor deviations in the information flow can be exponentially amplified during chaotic expansion, ultimately generating outputs that appear plausible but are detached from reality. The significant divergence of QLE in deeper layers provides evidence for this: when the model enters high-chaos regions, its information extraction paths become more susceptible to noise, leading to a surge in output space uncertainty.

These findings suggest that future research should incorporate "chaos control" mechanisms (e.g., dynamic regularization or attractor constraints) into model design to balance creative reasoning with logical reliability.